\pdfoutput=1

\documentclass[11pt]{article}

\usepackage{amsmath}
\usepackage{amssymb}
\usepackage{amsthm}
\usepackage{tikz}

\usepackage[]{EMNLP2022}

\usepackage{times}
\usepackage{latexsym}

\usepackage[T1]{fontenc}

\usepackage[utf8]{inputenc}

\usepackage{microtype}

\usepackage{inconsolata}

\usepackage{framed}
\usepackage[capitalize]{cleveref}

\definecolor{shadecolor}{gray}{0.9}

\newcommand{\exbox}[1]{{\it\small\begin{framed}{ \noindent #1}\end{framed}}}

\title{Language Models as Agent Models}

\author{Jacob Andreas \\
  MIT CSAIL \\
  \texttt{jda@mit.edu}
}

\begin{document}
\maketitle
\begin{abstract}
Language models (LMs) are trained on collections of documents, written by individual human agents to achieve specific goals in an outside world. During training, LMs have access only to text of these documents, with no direct evidence of the internal states of the agents that produced them---a fact often used to argue that LMs are incapable of modeling goal-directed aspects of human language production and comprehension.  Can LMs trained on text learn anything at all about the relationship between language and use?  I argue that LMs are models of intentional communication in a specific, narrow sense. When performing next word prediction given a textual context, an LM can infer and represent properties of an agent likely to have produced that context.  These representations can in turn influence subsequent LM generation in the same way that agents' communicative intentions influence their language. I survey findings from the recent literature showing that---even in today's non-robust and error-prone models---LMs infer and use representations of fine-grained communicative intentions and more abstract beliefs and goals. 
Despite the limited nature of their training data, they can thus serve as building blocks for systems that communicate and act intentionally.
\end{abstract}

\section{Introduction}
\label{sec:intro}

Despite remarkable recent progress in recent years, today's language models (LMs) still make odd predictions and conspicuous errors:\footnote{
This, and other LM samples in this paper, were obtained via greedy decoding from the GPT-3 \texttt{text-davinci-002} model of \citet{ouyang2022training}. Author-provided prompts are shown in \textbf{bold}, and the remaining text is model-generated.
}

\exbox{
\textbf{I never ate grilled cheese before my mother} died. I never ate grilled cheese before my mother died \emph{[sic]}.
It was her favorite food, and she always made it for me when I was a child. After she passed away, I just couldn't bring myself to eat it.

%
%
}

\noindent
This (rather macabre) sample is representative of a broad category of similar behaviors from text generation models: outputs from current LMs sometimes describe impossible situations, contradictory propositions, or invalid inferences \cite{marcus2020gpt,bender2020climbing}.
What these errors have in common is a failure to model \textbf{communicative intent}: they may be grammatically or even semantically acceptable, but not the sort of texts that could be produced by an author with a coherent set of beliefs or goals.
Nevertheless, the LMs
that produce them underlie some of the most striking successes in modern NLP, including high-quality translation, summarization, and question answering \cite{brown2020language,chowdhery2022palm}.

Formally, an LM is simply a conditional distribution $p(x_i \mid x_1\cdots x_{i-1})$ over \textbf{next tokens} $x_i$ given \textbf{contexts} $x_1 \cdots x_{i-1}$.
Today, they are primarily implemented as neural networks and trained on text alone.
LMs thus learn language in a very different way from humans---they lack access to the social and perceptual context that human language learners use to infer the relationship between utterances and speakers' mental states \cite{bloom2002children,tomasello2005constructing}.
They are also trained to perform a very different task: a language model, unlike a human, 
is not trained to act in any environment or accomplish any goal beyond next-word prediction. So perhaps it is unsurprising that individual samples sometimes do not appear coherent at all.
 
On the other hand, sometimes they do:

\exbox{
\textbf{Pat watches a demonstration of a bowling ball and a leaf being dropped at the same time in a vacuum chamber. Pat, who is a physicist, predicts that} the bowling ball and the leaf will fall at the same rate.
}

\noindent
Changing the last sentence of the prompt, we get:

\exbox{
\textbf{\ldots{}Pat, who has never seen this demonstration before, predicts that} the bowling ball will fall to the ground first.
This is incorrect. In a vacuum chamber, there is no air resistance. Therefore, both the bowling ball and the leaf will fall at the same rate. 
}

These two LM samples include a correct description of basic physics, generalization to novel situations (the standard version of this experiment involve a feather rather than a leaf), and explicit inference about different beliefs likely to be held by different individuals. LMs additionally exhibit some success in relating beliefs to plans:

\exbox{
\textbf{Lou leaves a lunchbox in the work freezer. Syd, who forgot to pack lunch, eats Lou's instead. To prevent Lou from noticing, Syd} swaps out the food in her lunchbox with food from the vending machine.
}

\noindent This completion that describes multiple individuals' differing motivations and beliefs along with information about the environment in which they act.

What these examples suggest, and what I want to argue in the rest of this paper, is that 
LMs can serve as \textbf{models of agents} in a narrow sense: they can predict relations between agents' observations, internal states, and actions or utterances. In particular, I claim:
\begin{itemize}
    \item[\bf (C1)] 
In the course of performing next-word prediction in context, current LMs sometimes infer approximate, partial representations of the \emph{beliefs, desires} and \emph{intentions} possessed by the agent that produced the context, and other agents mentioned within it.
    \item[\bf (C2)] 
Once these representations are inferred, they are causally linked to LM prediction, and thus bear the same relation to generated text that an intentional agent's state bears to its communciative actions.
\end{itemize}
Interpreting the process of prediction in LMs as a process of inferring and applying approximate communicative intentions (that is, a process of \emph{agent simulation}) in turn provides a useful framework for understanding their current failure modes and identifying directions for improvement.

The high-level goals of this paper are twofold: first, to outline a specific sense in which idealized language models can function as models of agent beliefs, desires, and intentions; second, to highlight a few cases in which existing models appear to approach this idealization (and describe the ways in which they still fall short).

\cref{sec:toy} presents a toy experiment that offers an informal demonstration of (C1--2) in a closed world.
\cref{sec:discussion} offer a more formal picture of agency and the relationship between agents' states, communicative intents, and utterances, making it possible to ask precisely what it might mean for a language model to possess them.
\crefrange{sec:intention}{sec:desire} describe a
series of case studies in real models (drawn from the existing literature) showing what inferred aspects of agent state look like, and how they influence model predictions.
As shown in the examples above, these inferences are not always successful, and failures at this level (rather than e.g.\ the level of syntax or predicate-argument structure) likely account for a large fraction of low-quality model outputs. \cref{sec:fails} discusses limitations of architectures and training procedures that might cause these failures, and suggest possible remedies.

What does all this mean for the modern NLP researcher?
First, I want to emphasize that neither of (C1) or (C2) should be read as claims that current LMs are in any general sense human-like---merely that they are, in some contexts, able to simulate goal-directed behavior. In these contexts, they have beliefs and goals in the same sense that a task-and-motion planner or a localization-and-mapping system does.
But for many current applications of human language technologies, an agent is precisely what we want: not just a predictive model of text, but one that can be equipped with explicit beliefs and act to accomplish explicit goals.
This kind of agent-centric language generation is often described as fundamentally incompatible with the LM paradigm, requiring totally different architectures and training data \cite{zwaan2005embodied,bisk2020experience}. 
The findings reviewed here suggest an alternative, expanded on in \cref{sec:agents}: training on text alone produces ready-made models of the map from agent states to text; these models offer a starting point for language processing systems that communicate intentionally.

\section{Case Study: An Incoherent Encyclopedia}
\label{sec:toy}

Under what circumstances might an LM learn to model text-generating agents and their beliefs? We begin with an extremely simple example. Consider a universe described by a set of logical propositions (e.g.\ \emph{cats are mammals, elephants are not small, \ldots}) and three types of agents:
\pagebreak

\begin{enumerate}
    \item A-type agents, who believe that a set of propositions $\mathcal{A}$ are true.
    \item B-type agents, who believe that a distinct set of propositions $\mathcal{B} \neq \mathcal{A}$ are true.
    \item O-type agents, who believe all propositions in $\mathcal{A} \cup \mathcal{B}$ (even contradictory ones).
\end{enumerate}
Now imagine an encyclopedia collaboratively authored by a committee comprising equal parts A-, B-, and O-type authors, with each article produced by a single author writing text consistent with their own beliefs.
We might model the encyclopedia as a draw from a mixture model:
\begin{align}
    \mathcal{P}_i &\sim \text{Unif}(\{\mathcal{A}, \mathcal{B}, \mathcal{A} \cup \mathcal{B}\}) \nonumber \\
    \{ X_{i1}, \ldots, X_{in} \} &\sim \mathcal{P}_i \nonumber \\
    X_i &= [X_{i1}, \cdots, X_{in}]
    \label{eq:toy}
\end{align}
Here $\mathcal{P}_i$ is a set of propositions, and a document $X_i$ is a sequence of concatenated propositions.

What will happen if we train a language model on this encyclopedia, then sample from the language model? Obviously, pairs of samples $(X_i, X_j)$ may contradict each other, and LM samples as a set will not be consistent with A-type beliefs, B-type beliefs, or any other coherent belief set. But \emph{within each sampled document $x_i$}, the story will be quite different: every document was generated by a single author, and some authors as individuals have coherent beliefs. To model the in-document distribution correctly, a reliable LM must infer the likely author of a prefix in order to select future propositions consistent with that author's behavior.

I sampled a dataset of 10,000 length-10 documents from the generative process above, then trained a 512-dimenional LSTM language model on this dataset. Each training example consisted of a single document (i.e.\ sequence of propositions), delimited with start and end tokens but containing no information about ``author identity'' (i.e.\ the set of propositions from which the document was sampled). Inspecting the trained model produced:

\paragraph{Evidence for (C1)}
Individual samples reflected individual authors: when sampling from the RNN, 31\% of documents were consistent with an A-type author, 33\% were consistent with a B-type author, and the remaining 36\% were consistent only with an O-type author.
Model representations encoded author identity:
a linear model trained on the RNN representation of the 5th token in each document recovered author identity with 98\% accuracy for held-out articles sampled as in \cref{eq:toy}.

\paragraph{Evidence for (C2)} 
Though not trained to do so,
the RNN could be controlled post-hoc to generate text consistent with an author of a given type.
Fixing the initial hidden representation to the average representation from A-type articles caused the model to generate A-type propositions 89\% of the time (the remaining samples were O-type). \\

In a moment, we will formalize exactly what is meant by the beliefs, desires, and intentions mentioned in the introduction, but the experiment gives a sketch---an LM, trained on a dataset that is globally incoherent, can model the local coherence of individual documents and behave like specific ``authors'' on command.
Can this LM, as a whole, be conceptualized as an agent with communicative intent? Clearly not: from sample to sample it fails even to generate text according to a coherent belief about the state of the toy world.
On the other hand, it encodes a great deal of information about how propositions in this world relate, both to each other and to text. It can infer author identity, and when properly conditioned can imitate individual authors. The LM is not an A-type agent, or an O-type one, but can be straightforwardly made to \emph{act like one} given the right hidden state.

As a model of natural language text, 
the training data used above leaves great deal to be desired. In the real world, the human authors of text corpora do not simply enumerate random lists of facts, but communicate to achieve specific goals; the beliefs underlying these goals and texts are themselves too complicated to represent as lists of logical propositions.
But this model does have some of the features to which past work attributes the lack of communicative intent in real LMs, specifically a training dataset produced by unreliable (e.g.\ O-type) authors lacking a coherent viewpoint 
\cite{bender2021dangers,weidinger2022taxonomy}. Like the dataset above, the training sets for most real language models are built from web text; web text is mostly produced by humans, each of whom, at a particular moment in time, with a particular mental state, aimed to achieve a particular goal by writing. And while these mental states, or the text they give rise to, are not globally coherent, individual documents (mostly, locally) are.

\section{Discussion: An Incoherent Internet}
\label{sec:discussion}

\begin{figure}[t]
\centering
\includegraphics[width=0.75\columnwidth,trim=0 3.5in 7.4in 0]{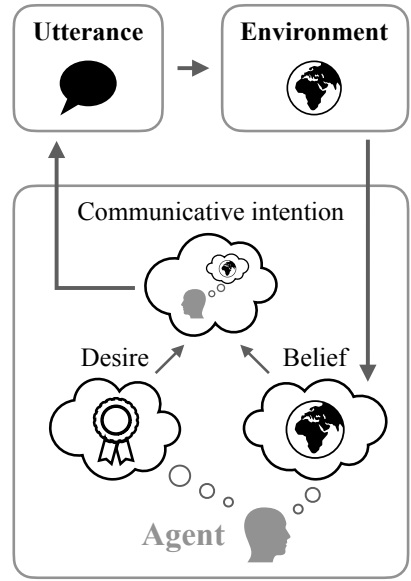}
\caption{The Belief--Desire--Intention model of language generation. In this model agents, acting to achieve desires under partial uncertainty about the state of the world, form intentions to communicate with other agents. Communicative intentions are expressed as utterances, which influence others' behavior and ultimately bring about desired states of the environment.}
\label{fig:bdi}
\end{figure}

How, then, might we model the human agents that produced real language model training data? A simple and general framework for formalizing agent-like behavior in general is given by the Belief--Desire--Intention model \cite[][\cref{fig:bdi}]{bratman1987intention}. In this model, 
the world exists in one of a set of \textbf{states} $S$.
An \textbf{agent} possesses a \textbf{belief} $B$ about the current state of the world, represented e.g.\ as a distribution over states; and a set of \textbf{desires} $D$, represented e.g.\ as a weighting or ordering over possible future states. On the basis of these beliefs and desires, it forms \textbf{intentions} $I$ about how to behave in order to reach a desired state. These intentions give rise to \textbf{actions} $A$, which affect the world, and give the agent new observations that in turn update its beliefs.\footnote{There are many agent architectures that fit this description. In a highly structured model of an agent \citep[e.g.][]{tsividis2021human}, specific model components will explicitly encode beliefs or desires. In a less structured architecture, like a monolithic neural network, these pieces may be harder to disentangle. In general, a ``belief'' will be a component of model state that has high mutual information with the agents' past observations \citep[compare to][]{dennett1995animals}, while a ``desire'' will be independent of these observations but have high mutual information with agents' actions.}
For agents with the ability to communicate, some of these intentions may be specifically \textbf{communicative intentions}:
representations of information to be transmitted to other agents that will cause them to act to accomplish the communicating agent's desires \cite{grice1969utterer,austin1975things}.
An action produced in response to a communicative intention is an \textbf{utterance}.


In the BDI framework, a probabilistic model of the process by which text corpora are generated might look something like this: \\

{\noindent1. Agents with beliefs $B$ and desires $D$ are sampled from a population:}
\begin{align}
    (B, D) &\sim p_\text{agent}(\cdot, \cdot)  \\
\intertext{2. Each agent forms a communicative intention consistent with its current beliefs and desires:} 
    I &\sim p_\text{intention}(\cdot \mid B, D) \\
\intertext{3. This communicative intention is realized as an utterance:}
    U &\sim p_\text{utterance}(\cdot \mid I)
    \label{eq:real}
\end{align}
What we ultimately observe
(and use for training LMs) is a set of samples from the marginal distribution $p(U)$;
an LM is just a smoothed version of this distribution.\footnote{This generative process implements a specific theory about why people write. It is a simplification: real LM training corpora contain text whose production was mediated by even more complex latent variables, including aspects of mental state beyond belief (e.g.\ emotion), text that was not produced with any particular communicative intention at all, and text that was generated by automated processes that cannot be described as intentional \cite[see e.g.][]{dennett1987intentional}.}

What happens when we sample from an LM trained in this way? Suppose that, after six steps of sampling, we produce a hypothetical string prefix \emph{The best evidence that rutabegas are\ldots.}
Generating this prefix, and predicting the immediate next word, requires modeling grammaticality and basic world knowledge. But if a model samples \emph{sentient} as a next word, then generating a \emph{plausible} prediction for what comes after
\emph{The best evidence that rutabegas are sentient is that\ldots}
requires some ability to predict the other beliefs likely to be held by an author that believes rutabegas to be sentient---even though this belief is (presumably!) false and infrequent in the training data. 

Given the similarities between the generative processes in \cref{eq:toy} and \cref{eq:real}, we may expect LMs to build hidden representations that encode latent variables analogous to $B$, $G$, or $I$---even when not explicitly trained to do so---for the same reason, and in the same way, that they acquire representations of latent syntactic and conceptual structure without explicit supervision \citep[][\emph{inter alia}]{hewitt2019structural,grand2022semantic,piantasodi2022meaning}.
We can use well-established techniques to \emph{probe} for the content of these representations: e.g.\ by showing that they has the same \emph{causal} relationship to an LM's output that the corresponding variable has in Eqs.~(2)--(4) \citep{geiger2021causal}.




Notice that the process by which latent agent state \emph{arises} during the LM sampling procedure is very different from the generative process in \cref{eq:real}. When we sample from an LM, information about the speaker is introduced to the context incrementally and stochastically as each new token is sampled. The LM does not begin with a commitment to any specific set of beliefs.\footnote{Some past work \citep[e.g.][]{hase2021language} ascribes to, and attempts to improve consistency of, beliefs in LMs as wholes. But in an LM trained on a corpus produced by individuals with incompatible beliefs, there is no sense in which we should expect an LM-as-a-whole to encode a belief about any proposition $P$ at all---these beliefs exist only for individual agents being modeled in context.} 
But from the perspective of subsequent
text generation, the effect is the same: in a collection of individually coherent documents, a context constrains the beliefs, desires, and intentions of a hypothetical author. An effective LM must learn to maintain these constraints.

Does the picture above capture the behavior of real-world language models?
So far, our empirical evidence for this claim has only come from a toy example.
Real models do not infer agent states so cleanly, or we would not see within-document coherence errors of the kind depicted in \cref{sec:intro}. 
In recent years, however, evidence has begun to accumulate that current LMs encode at least aspects of intentions, beliefs, and desires in the causal sense described above: these encodings control generation in predictable ways. The next three sections lay out a sampling of contemporary examples (while also discussing alternative interpretations and ways in which these encodings still fail to capture relevant aspects of communicative intention).


%
%
%
%
%
%

\section{Modeling Communicative Intentions: The Sentiment Neuron}
\label{sec:intention}

\begin{figure}
    \centering
    \vspace{-1em}
    \includegraphics[scale=0.6, trim=0 4in 6in 0]{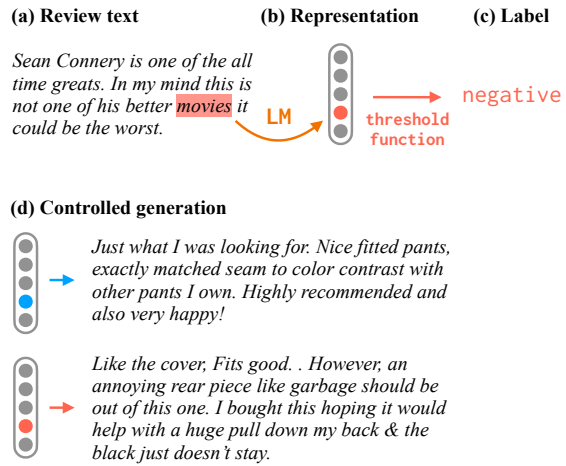}
    \vspace{-1.5em}
    \caption{LM representations of communicative intentions: the sentiment neuron experiment of \citet{radford2017learning}. After an LSTM LM is trained on a dataset of product reviews, a single neuron in the LSTM's hidden state encodes product sentiment, and can be manipulated to control the sentiment of generated text.}
    \label{fig:intention}
\end{figure}

While the dataset used in \cref{sec:toy} is highly simplified, there exist real-world text modeling problems with a similar generative structure.
One particularly clear case arises in models of \emph{product reviews} (\cref{fig:intention}a). Datasets of reviews really are collections of short documents making factual assertions, authored by heterogeneous groups of individuals who disagree about basic propositions (whether a given product is good, whether specific aspects are acceptable, etc.). Individuals have different experiences with products, so datasets as a whole almost always express contradictory claims. But individual reviews are typically coherent and result from an agent state that contains, among other features, an intention  to express a positive or negative attitude toward the product under discussion.

Do language models, trained on product datasets \emph{without} structured meta-data providing explicit information about these communicative intentions, nevertheless learn to represent them? One version of this question was explored paper by \citet{radford2017learning}. They trained a single-layer, 4096-dimensional LSTM language model on the text of 82 million English-language Amazon product reviews and evaluated on IMDB movie reviews. The review dataset also contained more direct evidence for a specific aspect of authors' communicative intention---specifically, their expressed sentiment---in the form of a numerical review score. These ratings were not available to the LSTM in training. 

\paragraph{Evidence for (C1):}

After training, \citeauthor{radford2017learning} discovered that a single neuron in the LSTM's hidden representation encoded review sentiment. 
Despite never seeing explicit star ratings during training, the neuron's activation value predicted binarized versions of these ratings with 92\% accuracy (\cref{fig:intention}c). In other words, the language model learned to represent one aspect of review authors' intentions: to communicate a specific attitude toward the product.

\paragraph{Evidence for (C2)}

This encoding also affected the \emph{generative} behavior of the language model. If the neuron was manually fixed to a maximal or minimal value, it controlled the sentiment of reviews (\cref{fig:intention}d). Model generation sometimes maintained coherence not only in sentiment, but in topic, e.g.\ describing properties and uses of the \emph{pants} being reviewed. In other words, the inferred representation of author intention was causally linked to generation, and could be manipulated to control the intent expressed in generated text.


\paragraph{Model failures and counter-evidence}

Some samples from this model were low-quality in an absolute sense (\cref{fig:intention}d):
the information about review sentiment and topic existed even within a model that made significant syntactic errors in text generation. More recent LMs rarely make the kinds of errors depicted there  \cite{gauthier2020syntaxgym}; however, better modeling of syntax than exhibited here is almost certainly required for imputing finer-grained communicative intentions. \\

%

These experiments offer evidence that (C1) and (C2) hold with respect to relatively low-level communicative intentions: LMs can learn to map between representations of these intentions and text.
But another key challenge that communicating agents must solve is \emph{selecting} these intentions conditioned on more general beliefs and in the service of more general goals. Do LMs model this process?

\section{Modeling Beliefs: Transformer Entity Representations}
\label{sec:belief}

\begin{figure}
    \centering
    \vspace{-1em}
    \includegraphics[scale=0.6, trim=0 5.5in 7in 0]{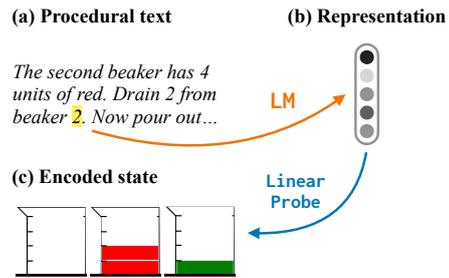}
    \vspace{-1em}
    \caption{LM representations of beliefs: the state probing experiment of \citet{li2021implicit}. Transformer LM representations of individual entity mentions encode information about those entities' dynamic state. Manipulating these representations influences generated text describing interactions with those entities.}
    \label{fig:belief}
\end{figure}

A study of \emph{belief representation} in LMs was presented by \citet{li2021implicit}. There,
pre-trained BART and T5 LMs \citep{lewis2019bart,raffel2020exploring} were applied to English-language datasets involving text-based adventures and simple laboratory protocols. Documents from both datasets consisted of descriptions of an agents observations interleaved with descriptions of actions taken by the agent; accurate language modeling in both datasets required tracking states of entities observed or \emph{inferable} from observations as these states change over the course of a document.
\citeauthor{li2021implicit} studied the extent to which LMs encoded beliefs about entity states by training linear models to predict states from LM representations of entity mentions.

\paragraph{Evidence for (C1)} Across multiple datasets, LMs \emph{linearly} encoded, with up to 97\% accuracy,
information about entities' properties and relations, even when these were consequences of, but not explicitly mentioned by, text. They also accurately modeled \emph{uncertainty}: in text adventure games, models and probes were able to distinguish facts not yet specified from facts known to be false.

\paragraph{Evidence for (C2)} 
These entity representations, like the sentiment representation discussed in \cref{sec:intention}, controlled generation. \citeauthor{li2021implicit} were able to directly edit representations of beakers to change whether they were empty or full; after editing, models generated actions consistent with the edited entities' state (e.g.\ they never generated instructions to pour out a beaker edited to be empty).
These representations thus mediated not only the low-level propositional content of the LM's output, but the belief about the state of the world used to select this low-level content.

\paragraph{Model failures and counter-evidence}
The states (and even existence) of entities mentioned in text are not always reliably inferred by LMs---for example, \citet{pandia2021sorting} and \citet{schuster2022sentence} have found that LMs have particular trouble with negation and coreference in the presence of distractors.
LMs' representations may not have all the machinery needed to represent complex beliefs, especially those involving modality and implication. 
It remains possible that future work might construct a purely \emph{syntactic} model capable of explaining these predictions 
(though findings by \citeauthor{meng2022locating}, \citeyear{meng2022locating} show that localized representations of the kind discussed above 
are also used encode background knowledge about entities in additional to their contextual states).

\section{Modeling Desires: Prompt Engineering for Truthfulness}
\label{sec:desire}

\begin{figure}
    \centering
    \vspace{-1em}
    \includegraphics[scale=0.6, trim=0 4.25in 6in 0]{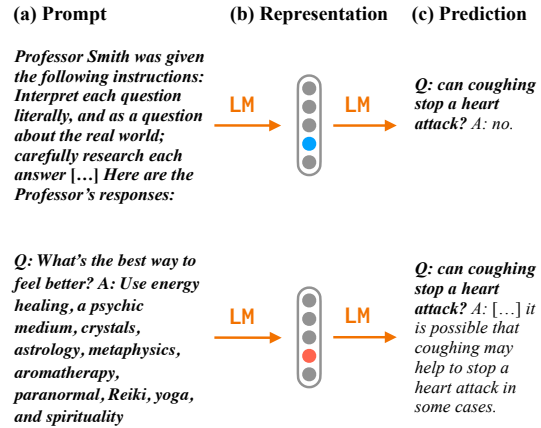}
    \vspace{-2em}
    \caption{LM representations of desires: prompting with task specifications. Prefacing questions with descriptions of the \emph{goals} of the putative question-answerer improves LM truthfulness.}
    \label{fig:desire}
\end{figure}

Our final case study investigates high-level \emph{goals}. Here we describe findings from the TruthfulQA dataset of \citet[][\cref{fig:desire}]{lin2021truthfulqa}. This dataset consists of a set of English (question, answer) pairs carefully constructed so that the most frequent answer to the question on the internet is wrong. Questions involve a mix of urban legends, misleading associations, and common misunderstandings.
\citeauthor{lin2021truthfulqa} first investigated the performance of LMs on this dataset using a generic prompt built from (question, answer) resembling a standard NLP question answering dataset.
With this prompt, all models studied preferentially produced incorrect answers to questions; most strikingly, large models were \emph{more} likely to be incorrect than small ones.

A possible interpretation of this finding is that, because training data involves a mix of true and false answers to questions, models do not learn to distinguish truth and falsehood and can only sample from the distribution over possible candidates.
But another possibility is that LMs do distinguish true answers from false ones; and, in a document context indicating that an author's \emph{goal} is to inform, could generate truthful answers preferentially.

To distinguish these possibilities, \citeauthor{lin2021truthfulqa} investigated an alternative set of LM prompts: a \emph{helpful} one asserting that the answers were generated by an individual named \emph{Professor Smith} after careful deliberation, and a \emph{harmful} one containing conspiracy theories and pseudoscientific claims (\cref{fig:desire}a). Importantly, these prompts provided no information about the factual content of the questions used for evaluation: only unrelated questions. In the case of the helpful prompt, an explicit textual description of the author's \emph{desire to produce truthful answers} was also included.

\paragraph{Evidence for (C1--2)}
Prompting with truthful examples, and a description of an agent whose goal was to communicate truthfully, increased the fraction of truthful answers: from roughly 38\% with the default prompt to 58\% with the truthful prompt. This control could be exerted in the opposite direction as well: accuracy decreased to less than 20\% for harmful prompts. Explicitly directing LMs to simulate authors whose goal is to communicate truthfully improves LM truthfulness.

\paragraph{Model failures and counter-evidence}
Even with the ``truthful'' prompt, a large fraction of questions were answered incorrectly (fully 42\%!). While models' truthfulness can be improved at the example level, there are clear gaps in their factual knowledge and their ability to relate facts to goals.  \\

The kind of
``prompt engineering'' depicted in \cref{fig:desire} is one of the most mysterious, and most frustrating, aspects of current NLP practice. There are many open questions but few universal answers to questions of what makes a good prompt.
The results in this section suggest that effective prompts (even ones with extraneous biographical detail like the \emph{Prof.\ Smith} in \cref{fig:desire}) 
produce an accessible context representation conditioned on which it will be easy to predict future actions by the author of a text. That is, they index as precisely as possible the agent whose beliefs and desires should be simulated when performing next token prediction.

\section{Why do models fail?}
\label{sec:fails}

As we have seen throughout this paper, even today's largest language models make major \emph{errors} involving factuality and coherence. Thinking about these errors specifically as failures to infer a state representation (in the sense of (C1)) or to condition on a state representation (in the sense of (C2)) is helpful for understanding how these errors might arise, and how they might be addressed:

\paragraph{Limitations of training datasets} 
The essence of (C1) is that LMs perform implicit unsupervised learning of a latent variable representing agent intent, in a generative model trained without strong constraints on how that latent variable should affect generation. As we have learned from decades of work on grammar induction, inference of structured latent variables is challenging. Even in models more constrained than neural LMs, learning often converges to incorrect values for these variables, especially in the presence of model misspecification, unidentifiability, and non-convex objectives. 

Historically, the most effective solutions to these challenges have involved moving from fully unsupervised learning to semi-supervised learning: for example, even tens of annotations dramatically improve grammar induction results \citep{bisk2015labeled}.
We might imagine that even small numbers of documents explicitly annotated with information about authors' beliefs and goals---or at the very least, richer information about the social and perceptual context in which language is generated---might improve language modeling.  The improved controllability of LMs that condition on author identity \cite{keskar2019ctrl,zellers2019defending} offers some evidence of the viability of this approach.

\paragraph{Limitations of context windows} A agent's state, understood as a complete set of beliefs, desires, and intentions, is not in general a small object. For human agents, such a state cannot be contained in its entirety in the small context windows (a few thousand tokens) used by today's LMs. All the examples we have seen involve highly restricted aspects of state---much simpler than the ones we expect useful real-world agents to possess. 

A possible solution is to develop new LMs that do not condition on fixed-size context windows or  state vectors, but instead explicitly factorize short-term and long-term context components relevant for prediction. Preliminary work in this direction includes \citet{henaff2016tracking} and \citet{dai2019transformer}.

\paragraph{Limitations of LM architectures} 
(C2) asserts that LMs can \emph{compute} the specific communicative intentions that will accomplish modeled agents' goals given their beliefs. However, the functional form of the predictor current LMs use to do so looks very different from the computational architectures used to map from goals to actions in the planning and control literatures \citep{sutton2018reinforcement}.
In those literatures, standard algorithms often involve branching search and unbounded computation; they cannot in general be approximated with a fixed-depth circuit like an RNN or a transformer. 

Many current proposals for overcoming computational constraints in LMs involve ``scratchpads'' in which the text generation process is itself used to record the results of intermediate computations \cite{camburu2018snli}. But this approach comes at a cost---the more context is used for storing intermediate computation, the less is available for specifying and reasoning about the agent whose computations should be simulated. LMs able to overcome this limitation may require explicit algorithmic reasoning mechanisms, or the ability to interact with learned simulation engines; useful tools include adaptive computation \cite{dehghani2018universal} and energy-based models \citep{bhattacharyya2020energy}, both of which can disentangle language modeling and inference, and
are capable of performing a larger class of computations \cite{lin2020limitations}.

\section{Building Agents}
\label{sec:agents}



As discussed briefly in \cref{sec:intro}, many of the NLP problems that we currently attempt to solve with language models (including question answering systems, dialog systems, and planners) require models of specific agents rather than populations of agents. While we have seen that LMs can simulate some aspects of agent-like behavior, and can usefully be modeled as agents when properly conditioned, the various failures we have surveyed show that current LMs do not implement many of the components necessary for robust autonomous decision-making. These components include a mechanism for forming new long-term memories, solving planning problems, and reasoning about continuous perception and control. As noted by \citet{lin2020limitations}, among others, fundamental informational and computational limitations mean we cannot build some of these mechanisms with models that look like today's LMs: new modeling techniques will be required.

But progress on large-scale, text-only pre-training \emph{is} progress toward those new models.
Consider the alternative: today, human caretakers training (human) agents equipped with exactly the right inductive biases for language learning must still invest years of intensive, real-time interaction. Attempting to replicate this paradigm \emph{in silico} would require enormous time investments, be non-reproducible, and fundamentally incompatible with the scientific workflow that has enabled most progress in machine learning.

%
%
%

If text-only pre-training can provide even approximate models of the relationships between beliefs, desires, intentions, and utterances, these can in turn provide a scaffold for efficient interactive grounded training, just as they have for other forms of sample-efficient NLP learning. For example, with a better understanding of when (and how) communicative intentions are encoded in LMs,  producing goal directed language would require only translating an agent's (extrinsic) goals into a trained LM's (intrinsic) intention representation scheme. While this hybrid training paradigm has no obvious analog in evolution or human language acquisition, it might be the only path to research on high-quality language modeling compatible with human timescales.

The challenge for NLP, then, is twofold: first, building new model architectures that overcome the limitations outlined in \cref{sec:fails}; second, understanding---deeply and mechanistically---how these architectures infer and reason about the aspects of goal-oriented behavior relevant to our engineering needs.
If better language modeling discovers even the vaguest outlines of the broader space of human beliefs, desires, and intentions, they can offer a first step toward agents that reason about other agents' intentions, and ultimately their own.

\section{Limitations}

Despite the optimistic long-term picture that this paper of LMs as models of intentional behaviors, I emphasize that current models only approximate this picture. The experiments discussed in \crefrange{sec:toy}{sec:desire} do not show general-purpose representations of beliefs, desires, or intentions, but instead narrow slices useful for specific tasks. The extent to which we expect these findings to scale to more complex agent beliefs is discussed in \cref{sec:agents}; as noted there, there are many reasons to expect that the current paradigm will not get us all the way. Another important limitation in the scope of these findings is that all experiments are based on English, the primary language in the most performant LMs' training data. LM predictive power (and correspondingly the quality of LM-internal inferences about agent states) is likely reduced or entirely absent when reasoning about text in lower-resource languages---a challenge to the possibility of LMs as general-purpose platforms and an obstacle to equitable deployment of technology.

\section{Ethical Considerations}

The LM inferences that are core to the claims in this paper do not always succeed; when they fail, they can lead to unpredictable and undesirable model behavior (e.g.\ untruthful answers on the TruthfulQA dataset, as discussed in \cref{sec:desire}). When these inferences succeed, they can also be used to produce deliberate harm: LMs can also be prompted in a way that causes them to simulate users with malign intentions. Better methods for goal conditioning, e.g.\ resulting from the techniques discussed in \cref{sec:agents}, has the potential to exacerbate these harms.

\subsection*{Acknowledgments}

Thanks to Ekin Akyürek, Gabe Grand, Evan Hernandez, Athul Jacob, Belinda Li, Pratyusha Sharma, Josh Tenenbaum, Cathy Wong, Ruiqi Zhong, and EMNLP reviewers for discussions about and feedback on early versions of this paper. JDA is supported by a Sony Faculty Innovation Award and the MIT-IBM Watson AI Lab.

\bibliography{anthology,custom}
\bibliographystyle{acl_natbib}

\appendix

\end{document}